\documentclass[letterpaper, 10 pt, conference]{ieeeconf}  
\usepackage{times}  
\usepackage{url}  
\usepackage{graphicx}  
\usepackage{multirow}  
\usepackage{amsmath}
\usepackage{algorithm}
\usepackage{textcomp}
\usepackage{mathtools}

\DeclarePairedDelimiter\floor{\lfloor}{\rfloor}
\usepackage[noend]{algpseudocode}
\usepackage{array}
\usepackage{cite}
\usepackage[keeplastbox]{flushend}
\newcolumntype{P}[1]{>{\centering\arraybackslash}p{#1}}
\newcolumntype{L}[1]{>{\raggedright\let\newline\\\arraybackslash\hspace{0pt}}m{#1}}
\makeatletter
\def\BState{\State\hskip-\ALG@thistlm}
\makeatother
\usepackage{xcolor}

\usepackage{amsmath}

\IEEEoverridecommandlockouts                              

\title{\LARGE \bf
Augmenting Knowledge through Statistical,\\Goal-oriented Human-Robot Dialog
}

\author{Saeid Amiri$^{1}$, Sujay Bajracharya$^{2}$, Cihangir Goktolga$^{1}$, Jesse Thomason$^{3}$, and Shiqi Zhang$^{1}$
\thanks{$^{1}$Amiri, Goktolga, and Zhang are with SUNY Binghamton; {\tt\small samiri1@binghamton.edu}} 
\thanks{$^{2}$Bajracharya is with Cleveland State University}
\thanks{$^{3}$Thomason is with the University of Washington}%
}

\begin{document}
\maketitle
\thispagestyle{empty}
\pagestyle{empty}

\begin{abstract}

Some robots can interact with humans using natural language, and identify service requests through human-robot dialog. 
However, few robots are able to improve their language capabilities from this experience. 
In this paper, we develop a dialog agent for robots that is able to interpret user commands using a semantic parser, while asking clarification questions using a probabilistic dialog manager.  
This dialog agent is able to augment its knowledge base and improve its language capabilities by learning from dialog experiences, e.g., adding new entities and learning new ways of referring to existing entities. 
We have extensively evaluated our dialog system in simulation as well as with human participants through MTurk and real-robot platforms. 
We demonstrate that our dialog agent performs better in efficiency and accuracy in comparison to baseline learning agents. Demo video can be found at \url{https://youtu.be/DFB3jbHBqYE} 

\end{abstract}

\section{Introduction}
\label{sec:intro}

Mobile robots have been extensively used to conduct tasks, such as guidance and object delivery, in the real world. Notable examples include the Amazon warehouse robots and the Relay robots from Savioke. However, these robots either work in human-forbidden environments, or have no interaction with humans except for obstacle avoidance. 
Researchers are developing mobile robot platforms that are able to interact with people in everyday, human-inhabited environments~\cite{khandelwal2017bwibots,hawes2017strands,chen2017robots,veloso2018increasingly}. 
Some of the robot platforms can learn from the experience of human-robot interaction (HRI) to improve their language skills, e.g., learning new synonyms~\cite{thomason2015learning}, but none of them learn entirely new entities. 
This work aims at a multitask dialog management problem, where a robot simultaneously identifies service requests through human-robot dialog and learns new entities from this experience to augment its internal knowledge base (KB). 

A robot dialog system typically includes at least three components for language understanding: state tracking, dialog management, and language synthesis. 
Our dialog agent includes the four components by further supporting dialog-based knowledge augmentation. 
Our dialog system is \emph{goal-oriented}, and aims at maximizing information gain.
In this setting, people prefer dialog agents that are able to accurately identify human intention using fewer dialog turns.


\begin{figure}[t]
  \begin{center}
    \vspace{.5em}
    \includegraphics[width=.8\columnwidth]{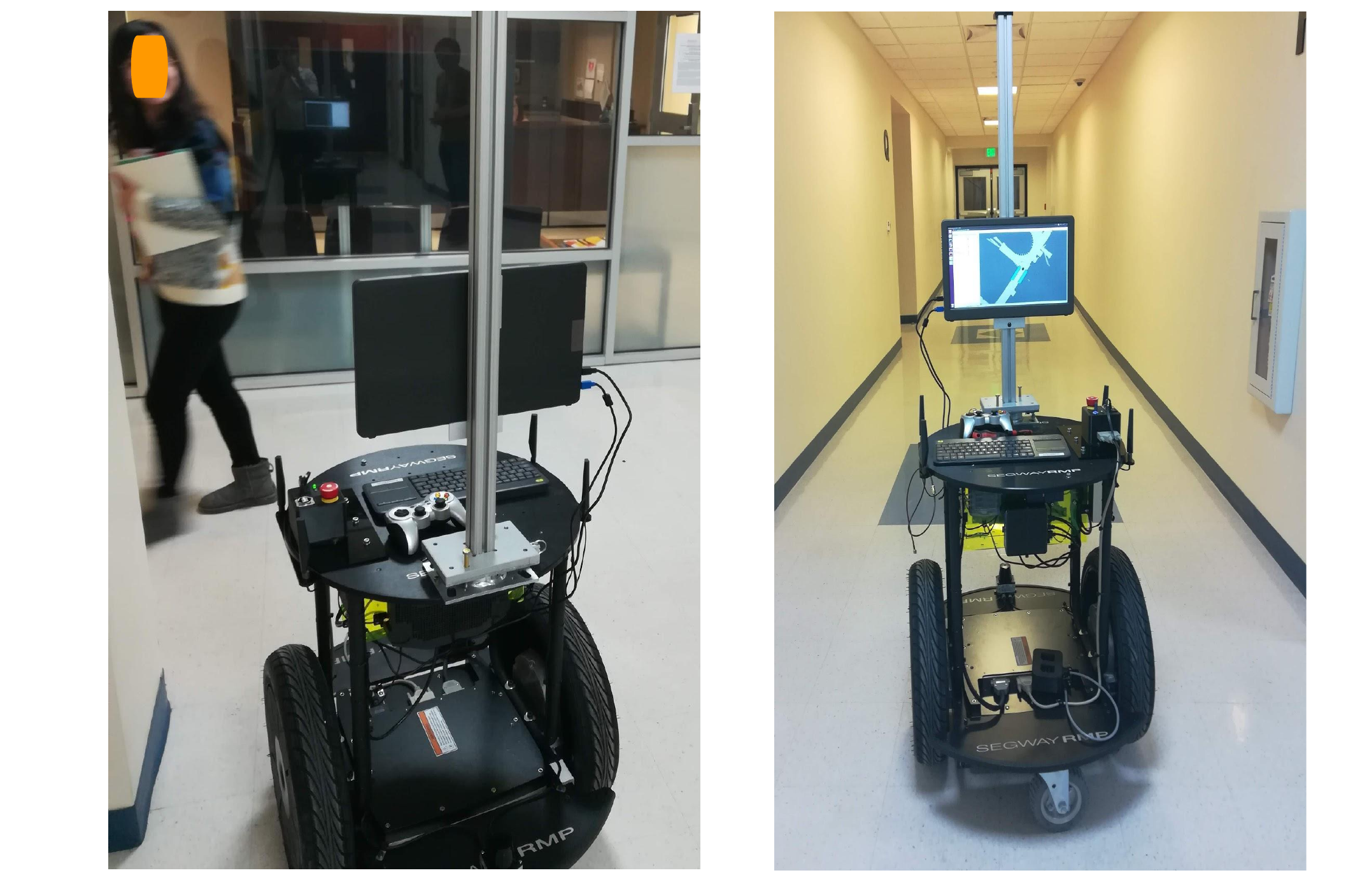}
    \vspace{-.5em}
    \caption{Our dialog agent is implemented and deployed on a Segway-based mobile robot platform (front and back). }
    \label{fig:robot}
  \end{center}
  \vspace{-1em}
\end{figure}

Goal-oriented dialog systems are necessary for language-based human-robot interaction because, in most cases, people cannot \emph{fully} and \emph{accurately} deliver information using a single dialog turn.
Consider a service request of ``\emph{Robot, please deliver a coffee to the conference room!}''
It is possible that the robot does not know which conference room the speaker is referring to, in which case it is necessary to ask clarification questions such as ``\emph{Where should I deliver a coffee?}'' in order to perform the correct action. To further identify the service request, the robot might want to ask about the recipient as well: ``\emph{For whom is the delivery?}''
Although such goal-oriented dialog systems have been implemented on robots, few of them can learn to improve their language capabilities or augment their KB from the experience of human-robot conversations in the real world (details in Section~\ref{sec:related}).\footnote{In comparison, there are dialog agents that aim at maximizing social engagement and prefer extended conversations, e.g., Microsoft XiaoIce, which are beyond the scope of this work.}

This work focuses on dialog-based robot knowledge augmentation, where the agent must identify \emph{when} it is necessary to augment its KB and \emph{where} in the KB to do that, as applied to our Segway-based mobile robot shown in Fig.~\ref{fig:robot}. 
In this paper, we develop a dual-track dialog manager to help the agent maintain a confidence level of how well the current dialog being supported by the KB, and accordingly decide to whether to augment its KB or not. 
After the agent becomes confident that new entities are necessary so as to make progress in the dialog, it decides where in the KB to add a new entity (e.g., a new \emph{item} or a new \emph{person} is being referred to by the user) by analyzing the flow of the dialog. 
As a result, our dialog agent is able to decide both \emph{when} and \emph{how} to augment its KB in a semantically meaningful way. 

Our dialog system has been evaluated in simulation and in the real world. 
Results show that our dialog system performs better in service request identification (both efficiency and accuracy), in comparison to baselines that use predefined strategies. 
Human-subject experiments suggest that our knowledge augmentation component improves user experience as well.



\section{Related Work}
\label{sec:related}

Researchers have developed algorithms for learning to interpret natural language commands~\cite{matuszek2013learning,misra2016tell,tellex2011understanding}.  
Recent research enabled the co-learning of syntax and semantics of spatial language~\cite{spranger2015co,gong2018temporal}. 
Although the systems support the learning of language skills, they do not have a dialog management component (implicitly assuming perfect language understanding), and hence do not readily support multi-turn communications.






Algorithms have been developed for dialog policy learning~\cite{singh2002optimizing,puterman2014markov,sutton1998reinforcement}. Recent research on Deep RL has enabled dialog agents to learn complex representations for dialog management~\cite{cuayahuitl2017simpleds,lu2019goal}. 
The systems do not include a language parsing component. 
As a result, users can only communicate with their dialog agents using simple or predefined language patterns.







Mobile robot platforms have been equipped with semantic parsing and dialog management capabilities. After a task is identified in dialog, these robots are able to conduct service tasks using a task planner~\cite{chen2017robots,zhang2015corpp,lu2017leveraging}. 
Although these works enable a robot to identify human requests via dialog, they do not enable learning from these experiences. 





Dialog agents for mobile service robots have been developed for identifying service tasks such as human guidance and object delivery~\cite{thomason2015learning,padmakumar2017integrated}. 
A dialog manager suggests language actions for asking clarification questions, and the agent is able to learn from human-robot conversations. 
These methods focus on learning to improve an agent's language capabilities, but do not augment its knowledge base in this process (i.e., only pre-defined people, objects, and environmental locations can be reasoned about by the robot). 
This work builds on the dialog agent implemented by \cite{thomason2015learning}, and introduces a dual-track dialog-knowledge manager and a strategy for augmenting the robot's knowledge base.

\begin{figure}[t]
  \begin{center}
    \vspace{.5em}
    \includegraphics[width=\columnwidth]{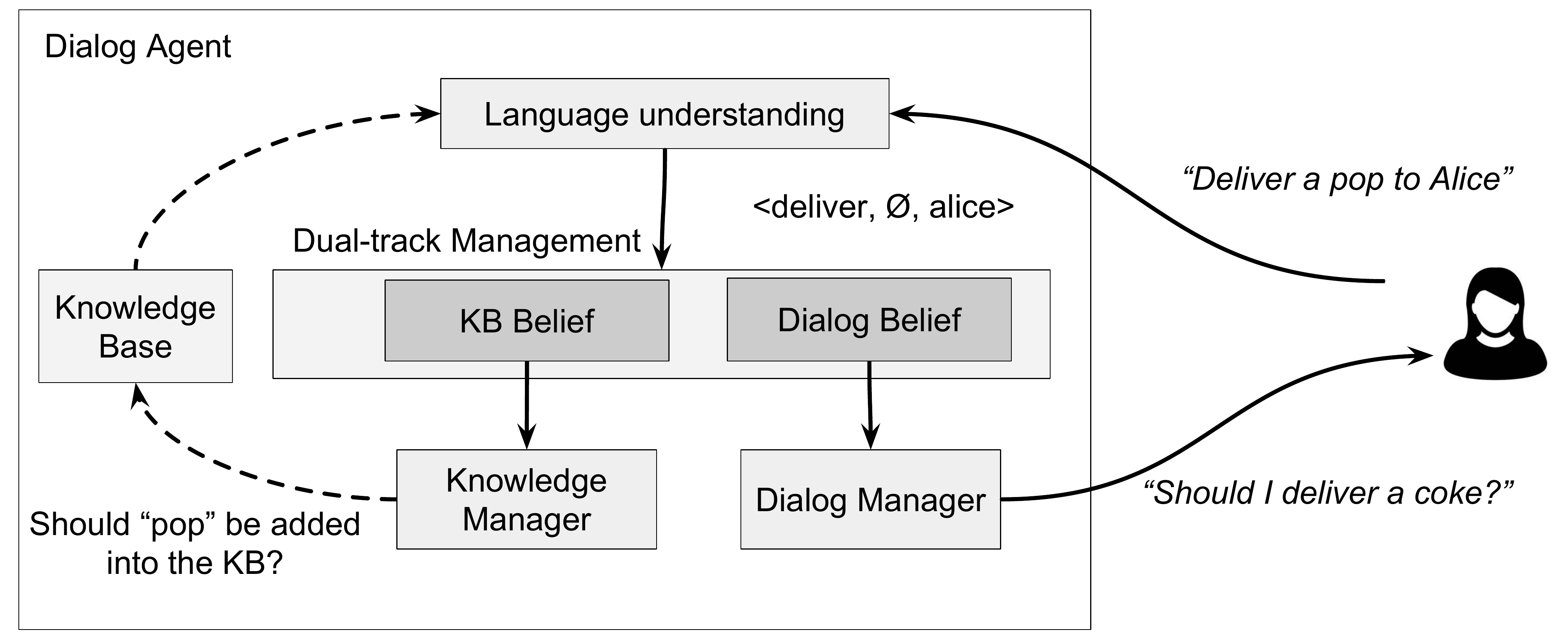}
    \vspace{-1.5em}
    \caption{A pictorial overview of our dialog system, including a hybrid parser for language understanding, and two management tracks for knowledge base (KB) and dialog respectively.} 
    \label{fig:overview}
  \end{center}
  \vspace{-1em}
\end{figure}




There are other dialog agents that specifically aim at knowledge augmentation through human-robot dialog~\cite{mericcli2014interactive,perera2015learning}. 
An instructable agent is able to learn new concepts and new procedural knowledge through human-robot dialog~\cite{azaria2016instructable}. 
Recent work enabled a mobile robot to ground new concepts using visual-linguistic observations, e.g., to ground new word ``box'' given a command of ``move to the box'' by exploring the environment and hypothesizing potential new objects~\cite{tucker2017learning}. 
These agents are able to augment their knowledge bases through language-based interactions with humans. However, their dialog management components (if any) do not model the noise in language understanding. 
Researchers developed a robot dialog system that focuses on situated verb semantics learning~\cite{she2017interactive}. Their dialog agent uses RL to learn a dialog management policy, and uses a semantic parser to process natural language inputs. 
A recent paper surveyed research on robot learning new tasks through natural language and action demonstration~\cite{chai2018language}.  
These works focused on learning the semantics of verbs, limiting the applicability of their knowledge augmentation approach. A recent work focuses on planning with open-world knowledge by reasoning about hypothetical objects while we focus more on modelling the language uncertainty \cite{jiang2019open}.




Our dialog agent is the first that together: 1) processes language inputs using a semantic parser to understand users' service requests, 2) leverages a dialog manager to account for the unreliability from the parser, and 3) augments its knowledge base using a knowledge manager.


\section{Dialog Agent}
\label{sec:agent}

In this section, we present our dialog agent that integrates a decision-theoretic dialog manager, and an information-theoretic knowledge manager, as illustrated in Fig.~\ref{fig:overview}. 


\subsection{Dialog and Knowledge Management}
Markov decision process (MDP) is a general sequential decision-making framework that can be used for planning under uncertainty~\cite{puterman2014markov}. 
Partially observable MDP (POMDP)~\cite{kaelbling1998planning} generalizes MDP to situations where ground truth world knowledge is fuzzy. 
POMDPs have been used for dialog management~\cite{young2013pomdp}, where the intentions of the interlocutors are latent.
There are two interleaved control loops in our dialog agent, resulting in a \emph{dual-track controller}. 
One track focuses on maintaining the dialog belief state, and suggests language actions to the agent. 
The other focuses on maintaining the belief of the current knowledge being (in)sufficient to complete the task, and suggests knowledge augmentation. 

Our dialog agent is implemented on a mobile service robot that communicates with human users using natural language to identify service tasks in the form of 
$$
    <task, item, recipient>
$$
where the agent must efficiently and accurately identify the service request (with unreliable language understanding capabilities) while augmenting KB on an as-needed basis. 


\vspace{.5em}
\subsubsection{Dialog Management Track}
The dialog management POMDP includes the following components: 

\begin{itemize}

\item $S: \{S^T \! \times \! S^I \! \times \! S^R\} \cup term $, where $S^T$ is the set of task \emph{types} (delivery and guidance in our case), $S^I$ is the set of \textit{items} used in the task, $S^R$ is the set of recipients of the delivery, and \textit{term} is the terminal state. 

\item $A: A^W \cup A^C \cup A^R$ is the action set. 
$A^W$ consists of general ``wh'' questions, such as ``\emph{Whom is this delivery for?}'' and ``\emph{What item should I deliver?}''. 
$A^C$ includes confirming questions that expect yes/no answers. Reporting actions $A^R$ return the estimated human requests.

\item $T: S \! \times \! A \! \times \! S' \! \rightarrow [0,1] $ is the 
state-transition function. In our case, the dialog remains in the same state after question-asking actions, and reporting actions lead transitions to \emph{term} deterministically. 

\item $R:S \times A \rightarrow {\rm I\!R}$ is the reward function. The reward values are assigned as: 
\[
    R(s,a)= 
\begin{cases}
    r^C,& \text{if } s\in S , a \in A^C \\
    r^W,& \text{if } s\in S , a \in A^W \\
    r^+,& \text{if } s\in S , a \in A^R , s \odot a \\
    r^-,& \text{if } s\in S , a \in A^R , s \otimes a 
\end{cases}
\]
where $r^C$ and $r^W$ are the costs of confirming and general questions, in the form of negative, relatively small values; $r^+$ is a big bonus for correct reports; and $r^-$ is a big penalty (negative) for incorrect reports.

\item $Z: Z^T \cup Z^I \cup Z^R \cup \{z^+ , z^-\}$ is the set of observations, where $Z^T$, $Z^I$ and $Z^R$ include observations of \textit{task type}, \textit{item}, and \textit{recipient} respectively. $z^+$ and $z^-$ correspond to ``\emph{yes}'' and ``
\emph{no}''.
Our dialog agent takes in observations as semantic parses that have correspondence to elements in $Z$. Other parses, including the malformed ones, produce random observations (detailed shortly). 

\item $O: S \times A \times Z \cup \textit{inapplicable}$ is the observation function that specifies the probability of observing $z \in Z$ in state $s \in S$, after taking action $a \in A$. 
Reporting actions yield the \emph{inapplicable} observations. 
Our observation function models the noise in language understanding, e.g., the probability of correctly recognizing $z^+$ (``\emph{yes}'') is $0.8$. 
The noise model is heuristically designed in this work, though it can be learned from real conversations. 
\end{itemize}

Solving this POMDP generates a policy $\pi$, which maps a belief to a language action ($a\in A$) that maximizes long-term information gain. 

\vspace{.5em}
\subsubsection{Knowledge Management Track}

In addition to the dialog management POMDP, we have a knowledge management POMDP that monitors whether the agent's knowledge is sufficient to support estimating human intentions. 
The knowledge management POMDP formulation is similar to that for dialog management but includes entities for unknown items and recipients.  
The components of the knowledge management POMDP are:

\begin{itemize}

\item $S^+ \! : \! S \cup 
\{ (s^T,s^I,\hat{s}^R) ~|~ \forall s^T \!\in\! S^T, \forall s^I \!\in\! S^I \}
\cup 
\{ (s^T, \hat{s}^I, s^R) ~|~ \forall s^T \!\in\! S^T, \forall s^R \!\in\! S^R \}$ 
the set of states. 
It includes all states in $S$ along with the states corresponding to new entities that correspond to an unknown item $\hat{s}^I$ and an unknown recipient $\hat{s}^R$;

\item $A^+: A \cup \{\hat{a}^I, \hat{a}^R \} \cup \hat{A}^R$ the set of actions including the actions in $A$, two actions ($\hat{a}^I$ and $\hat{a}^R$) for confirming the unknown \emph{item} and \emph{recipient}, and $\hat{A}^R$, reporting actions that correspond to the states in $S^+$;

\item $Z^+ = Z \cup \{\hat{z}^I, \hat{z}^R\}$ the augmented observation set, including $\hat{z}^I$ and $\hat{z}^R$ for unknown item and recipient. 
\end{itemize}

Transition and observation functions are generated accordingly and hence not listed. 

At runtime, we maintain belief distributions for both tracks of POMDPs. 
Belief $b$ of dialog POMDP is used for sequential decision making and dialog management, and 
belief $b^+$ of knowledge POMDP is only used for language augmentation purposes, i.e., determining when it is necessary to augment the KB.
When observations are made (observation $z \in Z$), both beliefs are updated using the Bayes update rule~\cite{kaelbling1998planning}. 
In our dialog system, observations are made based on the language understanding using a semantic parser.

The dual-track controller identifies the first contribution of this work. 
We use a dual-track, instead of merging them to unify the action selection process, because the knowledge track is only used for the purpose of maintaining beliefs (not for action selection) and modeling the uncertainty of unknown entities in a single controller will result in unnecessarily long dialogs. 
Separating the two tracks reduces the learning complexity of the entire framework.


\subsection{Language Understanding}

\begin{figure}[t]
  \begin{center}
    \vspace{.5em}
    \includegraphics[width=\columnwidth]{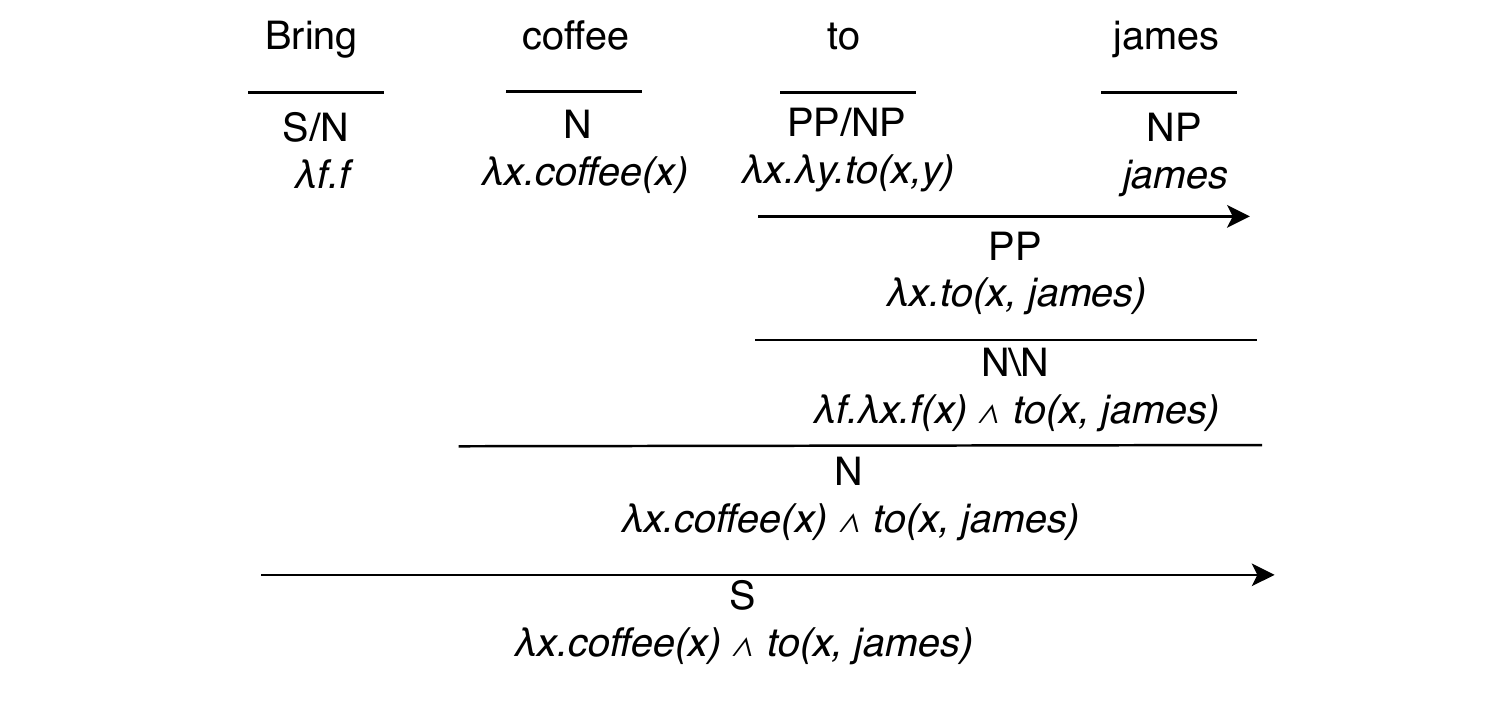}
    \vspace{-2em}
    \caption{An example of parsing a service request sentence using CCG semantic parsing and $\lambda$ calculus. }
    \label{fig:parse}
  \end{center}
  \vspace{-1em}
\end{figure}

In order to understand natural language and make observations for POMDPs, we use a semantic parser that builds on the Semantic Parsing Framework (SPF) described in~\cite{artzi2013cornell}.  
The input of the semantic parser is natural language from human users, and the output is a list of possible parses for a given sentence. 
Using the semantic parser, the request in natural language is transformed to a formal representation compatible with the robot's internal KB.  

Figure~\ref{fig:parse} shows an example of the parser recognizing a sentence.  
It can reason over the ontology of the known words when it parses a sentence, e.g., \emph{james:pe} and \emph{coffee:it}.  The dialog manager can use this information to translate from words to the corresponding observation for the question asked by the robot.  If the language understanding fails (e.g., producing parses that are malformed or do not comply with the preceding questions), then a random observation from $Z$ will be made for the unknown part of the request (introducing enough entropy to move the dialog along).


\subsection{Domain Knowledge Representation}
\label{sec:knowledge}

We use Answer Set Prolog (ASP)~\cite{gelfond2014knowledge}, a declarative language, for knowledge representation. 
The agent's knowledge base (KB), in the form of an ASP program, includes rules in the form of: 
$$
	l_0 \leftarrow l_1,~ \cdots,~ l_n, ~\textnormal{not}~ l_k,~ \cdots, ~\textnormal{not}~ l_{n+k}
$$
where $l$'s are literals that represent whether a statement is true. The right side of a rule is the \emph{body}, and the left side is the \emph{head}. The \texttt{not} symbol is called default negation, representing no evidence supporting a statement. 

The KB of our agent includes a set of \emph{entities} in ASP: 
\{\emph{alice, sandwich, kitchen, office1, delivery, }$\cdots$\}, where \emph{delivery} specifies the task type. 
A set of \emph{predicates}, such as 
\{\emph{recipient, item, task, room}\}, are used for specifying a category for each object.
As a result, we can use ASP rules
to fully specify tasks, such as ``\emph{deliver a coke to Alice}'': 
\begin{align*}
	&task(delivery).\\
	&item(coke).\\
	&recipient(alice).
\end{align*}

One can easily include more information into the ASP-based KB, such as rooms, positions of people, and a categorical tree of objects. Robot's KB is built on a lexicon that is a collection of information about the words of a language about the lexical categories. This ASP-based KB can be used for query responding and task planning purposes, where the query and/or task are specified by our dialog agent.

\begin{figure*}[t]
  \vspace{-1em}
  \begin{center}
    \hspace*{-1em}
    \includegraphics[width=1.\textwidth]{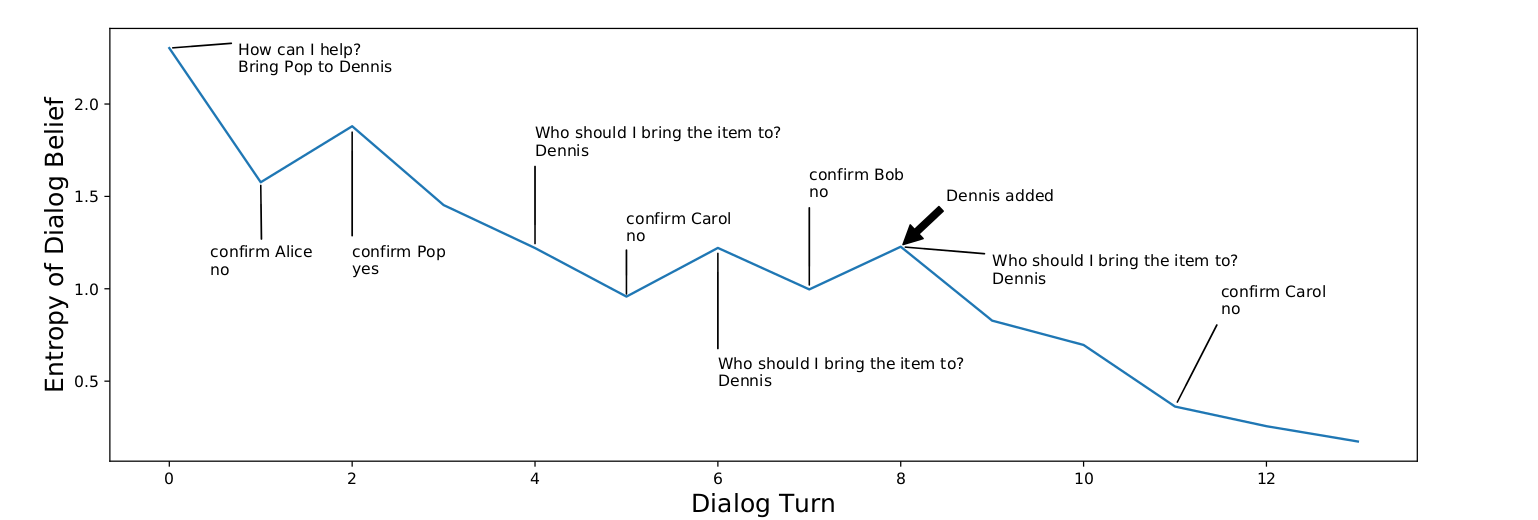}
    \vspace{-1em}
    \caption {In this example, the user requested a \emph{Pop} to a novel recipient, \emph{Dennis}.  The dialog agent understood \emph{Pop}, but not \emph{Dennis} (Turn 0), and it mistakenly observed \emph{Alice} as the recipient.  The user denied \emph{Alice} (Turn 1), and confirmed \emph{pop} (Turn~2). The conversation continued, as our dialog agent kept trying to identify the recipient, until the number of EFs crossed a threshold ($5$ in this case) in Turn 8. 
    Accordingly, \emph{Dennis} was added into the KB as a new recipient entity.  The agent continued asking clarification questions while being aware of \emph{Dennis}, until the dialog manager suggested the correct reporting action and delivers \emph{pop} to \emph{Dennis}. Although the clarification question may frustrate human, it makes the robot more confident in estimating human request. 
}
    \label{fig:ef}
  \end{center}
  \vspace{-1.5em}
\end{figure*}

\subsection{Algorithm for Knowledge Augmentation}
\label{sec:alg}


We define a few functions before introducing the main algorithm for simultaneous dialog management and knowledge augmentation. 
We use \emph{entropy} to measure the uncertainty level of the agent's belief distribution: 
$$
	H(b) = -\sum_{i=0}^{n-1} b(s_i) \cdot \log{b(s_i)} 
$$
When the agent is (un)confident about the state, the entropy value is (high) low. 
In particular, a uniform belief distribution corresponds to the highest entropy level. 
We use entropy for the following two purposes in our algorithm. \\
\vspace{-1em}
\paragraph*{I) Rewording Service Request} If the belief entropy is higher than threshold $h$ (meaning the agent is highly uncertain about the dialog state), we encourage the human user to state the entire request in one sentence. 
  Otherwise, the dialog manager decides the flow of the conversation.  
  
\vspace{1.em}
\paragraph*{II) Entropy Fluctuation} We introduce the concept of \emph{entropy fluctuation} (EF): 
\begin{align*}
      f(\textbf{b}) = &sign\Big(H(\textbf{b}[1]) - H(\textbf{b}[0])\Big) \oplus \\
      &sign\Big(H(\textbf{b}[2]) - H(\textbf{b}[1])\Big)  
\end{align*}
where $\textbf{b}$ is a belief queue that records dialog beliefs of the last three dialog turns, $f(\textbf{b})$ outputs \emph{true}, if there is an EF in the last three beliefs (i.e., entropy of the second last is the highest or lowest among the three), and $\oplus$ is the \emph{xor} operator.



\begin{algorithm}[t]\small
\caption{Dialog-based Knowledge Augmentation}
\label{alg}
\begin{algorithmic}[1]
\Require \text{$\mathcal{M}, \mathcal{M}^+, \tau_{b}, h, \Delta$, and a POMDP \emph{solver}}
\State \text{Initialize $b, b^+$ with uniform distributions}
\State Initialize EF counter $\delta \gets 0$ 
\State Initialize queue \textbf{b} of size 3 with $\{b,b,b\}$

\Repeat

\If {$~~\sum_{s^R=\hat{s}^R} b^+\big(s(s^T,~s^I,~s^R)\big) \!>\! \tau_{b}~~$}
  	\State \text{Add a new recipient entity in KB}
\ElsIf {$~~\sum_{s^I=\hat{s}^I} b^+\big(s(s^T,~s^I,~s^R)\big) \!>\! \tau_{b}~~$}    
  	\State \text{Add a new item entity in KB}
\ElsIf {$\delta \!>\! \Delta$ }
  	\State \text{Add (item or recipient) entity that is more likely}
\EndIf


\If {$f(\textbf{b})$ \textbf{is} \emph{true}}
	\State $\delta \gets \delta + 1 $
\EndIf
\If {$H(b) > h$}
	\State $a \gets$ ``\emph{Please reword your service request}''
\Else
    \State $a \gets \pi(b) $
\EndIf    
\State $o \gets parse(human~response) $ 
\State $\textnormal{Update}~b$ based on observation $o$ and action $a$
\State \textbf{b}.enqueue($b$) 
\If {$a \in A^C$}
\State $\textnormal{Update}~b^+$ based on observation $o$ and action $a$
\EndIf
\Until $s$ \textbf{is} $term$
\State \textbf{return} the request based on the last (reporting) action, and the (possibly updated) knowledge base. 
\end{algorithmic}
\end{algorithm}

\paragraph*{Algorithm for Dialog-Knowledge Management}
Algorithm~\ref{alg} shows the main operation loop of the dialog system. 
$\mathcal{M}$ and $\mathcal{M}^+$ are models for dialog-track and knowledge-track control respectively; $\tau_b$ is a probability threshold; $h$ is an entropy threshold; and $\Delta$ is a threshold over the number of EFs. 

The algorithm starts by initializing the two beliefs with uniform distributions. 
$\delta$, which counts the number of EFs, is initialized to $0$.  
If the marginal probability over $\hat{s}^R$ (or $\hat{s}^I$) of knowledge belief $b^+$ is higher than threshold $\tau_b$, or the number of EFs is higher than $\Delta$, we add a new entity into the KB. 
If the entropy of dialog belief is higher than $h$, then the agent asks for rewording the original service request. 
Otherwise, we use the dialog POMDP to maintain the dialog flow. 
Finally, the knowledge belief is only updated by confirming questions, which are able to invalidate agent hypothesis of unknown entities. 
The algorithm returns the request and the updated knowledge base.  
When adding a new entity, the agent explicitly asks the user to help specify the name of the unknown item (or person).  
The KB is updated along with the robot's lexicon for the semantic parser.  
The index for the unknown item or person is associated with the new entry. 
 We utilized two functions to calculate the parameters $\tau_b$ and $\Delta$: 

 $\tau_b (|KB|) =  \cfrac{1}{1 + e^-\floor*{\sqrt{|KB|}} } -  \cfrac{1}{\floor*{\sqrt{|KB|}}}  $

$\Delta(|KB|) = max(0,\floor*{\sqrt{|KB|}})  $ \\

With the new knowledge added to KB, POMDPs are dynamically constructed so that the dialog can continue seamlessly, and the belief $b$ is replaced with reinitialized $b^+$.
Fig.~\ref{fig:ef} illustrates an example dialog.


\section{Experiments}

\label{sec:experiments}
We have evaluated our dialog agent both in simulation and with human participants. 
When the user verbalizes the entire request, the agent receives a sequence of three (unreliable) observations on \emph{task}, \emph{item} and \emph{recipient} in a row. 
Unreliable language understanding is modeled in POMDP observations, e.g., the agent can correctly recognize ``coffee'' with probability ($0.8$ in our case), and this probability decreases given more items in the KB. 
The reward of confirming questions is $R^C\!=\!-1.0$, and the reward of wh-questions is $R^W\!=\!-1.5$. The above settings were shared in experiments both in simulation and with human participants. 
POMDPs are solved using an off-the-shelf system~\cite{kurniawati2008sarsop}.

Experiments were mainly designed to evaluate the following hypotheses: 
Our algorithm is able to 
I) Efficiently and accurately identify whether there is the need for KB augmentation or not, in case there is the need; 
II) Augment KB with higher F1 score under the noise in language understanding; and
III) Both augment KB and recognize human intention in the service request with higher success rate while minimizing QA cost. 

We compared our algorithm with two learning baselines that use predefined strategies to update their KB: Baseline-I augments KB only when the marginal probability of $\hat{s}^R$ ($\hat{s}^I$) reaches $\tau_b$, and Baseline-II augments KB only when number of EF $\delta$ reaches threshold $\Delta$.

Evaluation metrics used in the experiments consist of: 
\textbf{QA cost}, the total cost of QA actions; 
\textbf{Accuracy}, in an accurate trial, robot correctly identifies its KB entity inadequacy;
\textbf{Success rate}, where a trial is deemed successful, if the service request is correctly identified \emph{and} (if needed) the KB is correctly augmented; and 
\textbf{Dialog reward}, where QA cost and bonus/penalty are considered together. 
Focusing on the knowledge augmentation accuracy, we also use \textbf{F1 score} as a harmonic average of precision and recall in evaluation.

\subsection{Experiments in Simulation}
\begin{figure}[t]
  \begin{center}
    \vspace{.5em}
    \includegraphics[width=\columnwidth]{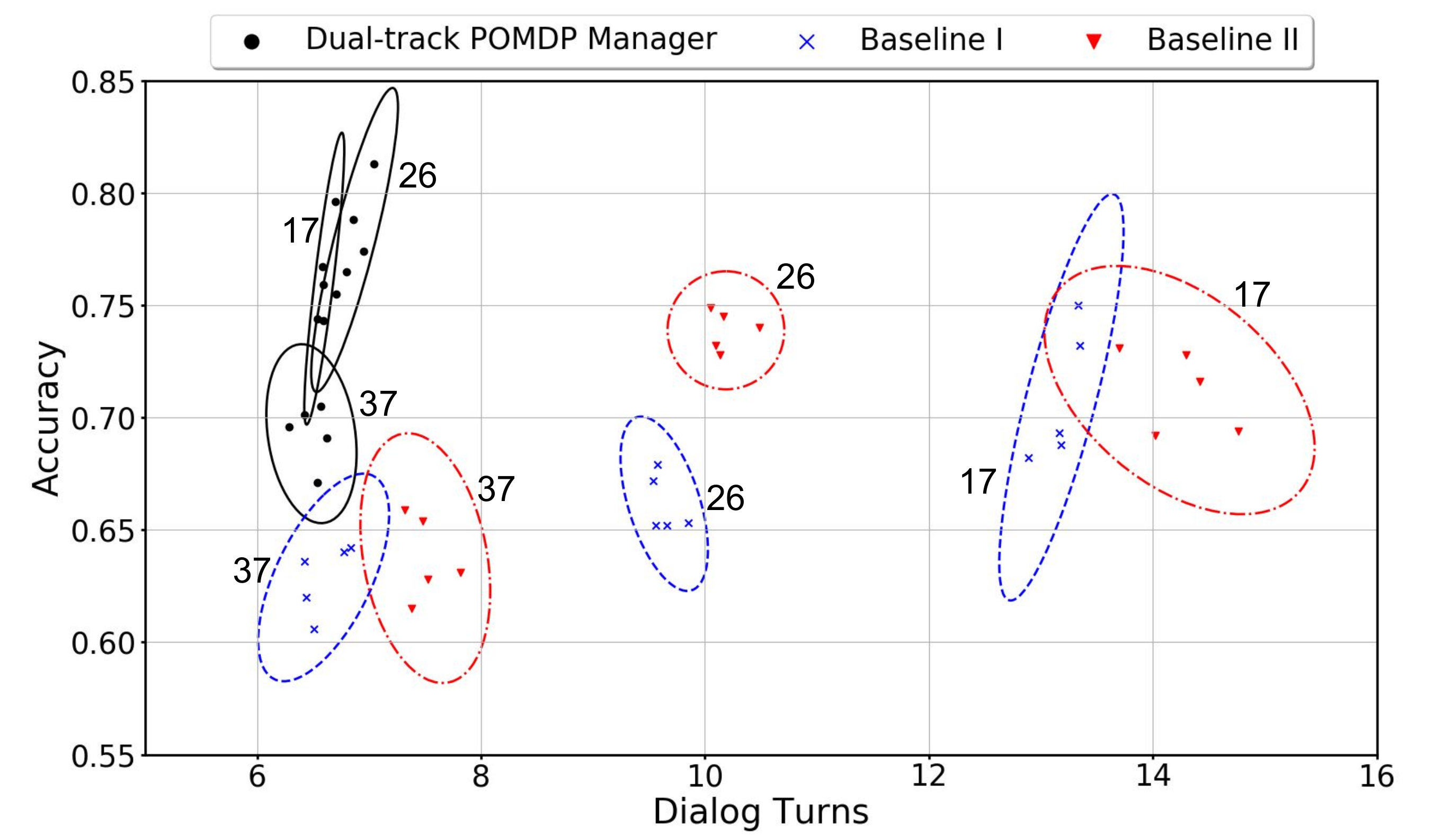}
    \vspace{-1em}
    \caption{Our dialog agent (Shown in $\bullet$) is able to detect the \emph{need} of KB augmentation with higher accuracy in fewer dialog turns compared to baselines. Covariance error ellipses calculated for 5 batches for each domain size. The numbers next to data points denotes the KB size.}
    \label{fig:hyp1}
  \end{center}
  \vspace{-1em}
\end{figure}





\begin{table}[t]\footnotesize
\caption{F1 Score of KB Update Given Different KB Sizes}
 \vspace{-1em}
\label{tab:3}
\begin{center}
\begin{tabular}{|c|c|c|}
 \hline
 \textbf{Agent} & \textbf{KB Size} & \textbf{F1 Score (std.)}  \\
 \hline \hline
 \emph{Dual Track Manager} &  & 0.79 (0.020) \\ \cline{1-1}\cline{3-3}
 \emph{Baseline I} & \multirow{2}{*}{17}  & 0.59 (0.030) \\
 \cline{1-1}\cline{3-3}
 \emph{Baseline II} &  & 0.61 (0.017) \\
 \hline \hline
 \emph{Dual Track Manager} &  & 0.77 (0.025) \\ \cline{1-1}\cline{3-3}

 \emph{Baseline I} & \multirow{2}{*}{26}  & 0.52 (0.016) \\
 \cline{1-1}\cline{3-3}
 \emph{Baseline II} &  & 0.66 (0.007) \\
 \hline \hline
 \emph{Dual Track Manager} &  & 0.62 (0.011) \\ \cline{1-1}\cline{3-3}
 
 \emph{Baseline I} & \multirow{2}{*}{37} & 0.47 (0.019) \\
\cline{1-1}\cline{3-3}
 \emph{Baseline II} &  & 0.46 (0.022) \\
 \hline
\end{tabular}
\end{center}
\vspace{-1.2em}
\end{table}

\begin{figure}[t]
  \vspace{.3em}
  \begin{center}
    \hspace*{0em}
    \includegraphics[width=\columnwidth]{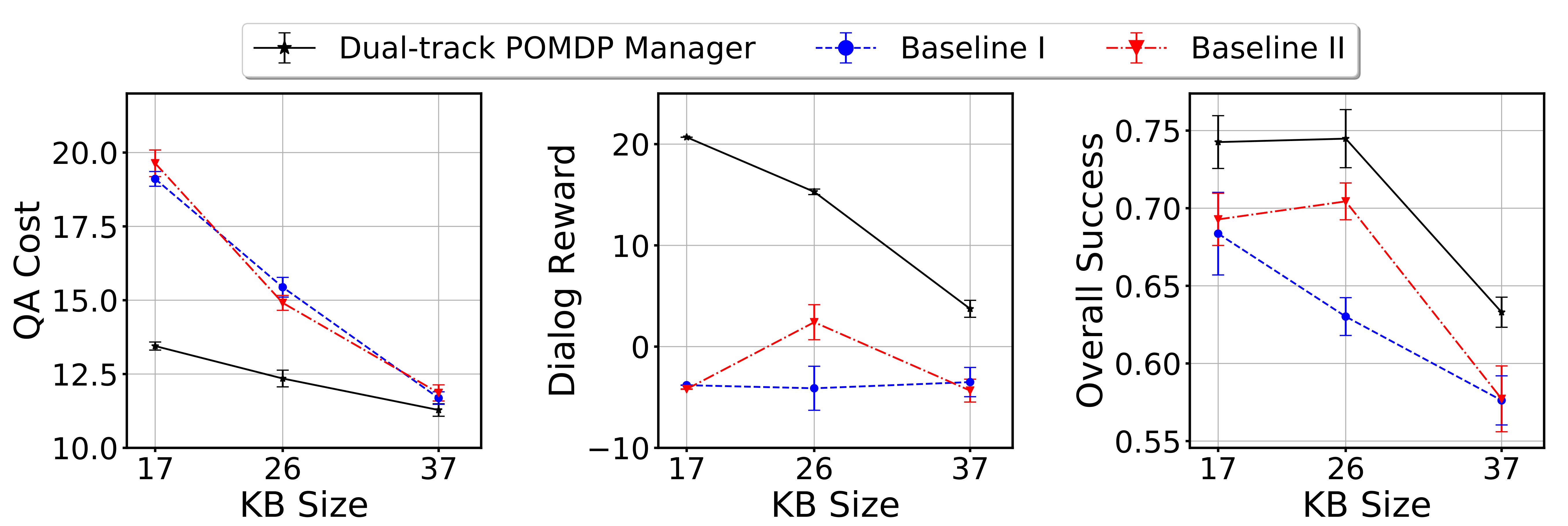}
    \hspace*{-1em}
    \vspace{-1.8em}
    \caption {Comparison between our agent and baselines in terms of both dialog and knowledge management. }

    \label{fig:hyp3}
  \end{center}
  \vspace{-1em}
\end{figure}

To evaluate each of the hypotheses, we simulated 3,000 trials over various domain sizes.
In each trial, a task, an item, and a recipient are sampled. 
\emph{$|KB|$} is all possible combinations of item/recipient plus the terminal state. For instance, $|KB|$ = 17 corresponds to a domain with 1 task, 4 items and 4 recipients. 
In the first experiment, we evaluated KB update accuracy versus the dialog turn in which the KB update has occurred (Hypothesis-I). 
Our algorithm consistently detects the need for KB augmentation earlier (fewer dialog turns) and with higher accuracy while baselines require longer conversations to figure out if they need a KB update (Fig.~\ref{fig:hyp1}).   
 
We further evaluated whether the entity added to KB matches with the human intention or not (Hypothesis-II). 
As presented in Table~\ref{tab:3}, our algorithm consistently maintains higher F1 score in comparison to other baseline agents in the medium sized KB. 
Finally, we evaluated how our algorithm is capable of both correct KB augmentation as well as correct execution of the task (Hypothesis-III). 
Figure \ref{fig:hyp3} shows that, our agent consistently maintains higher dialog reward while achieving lower QA cost and higher overall success. As the domain size increases, the agent gives up asking further questions that results in lower overall success and reward.

\subsection{Experiments with Human Participants}
\begin{figure}[t]
  \begin{center}
    \vspace{.5em}
    \includegraphics[width=0.95\columnwidth]{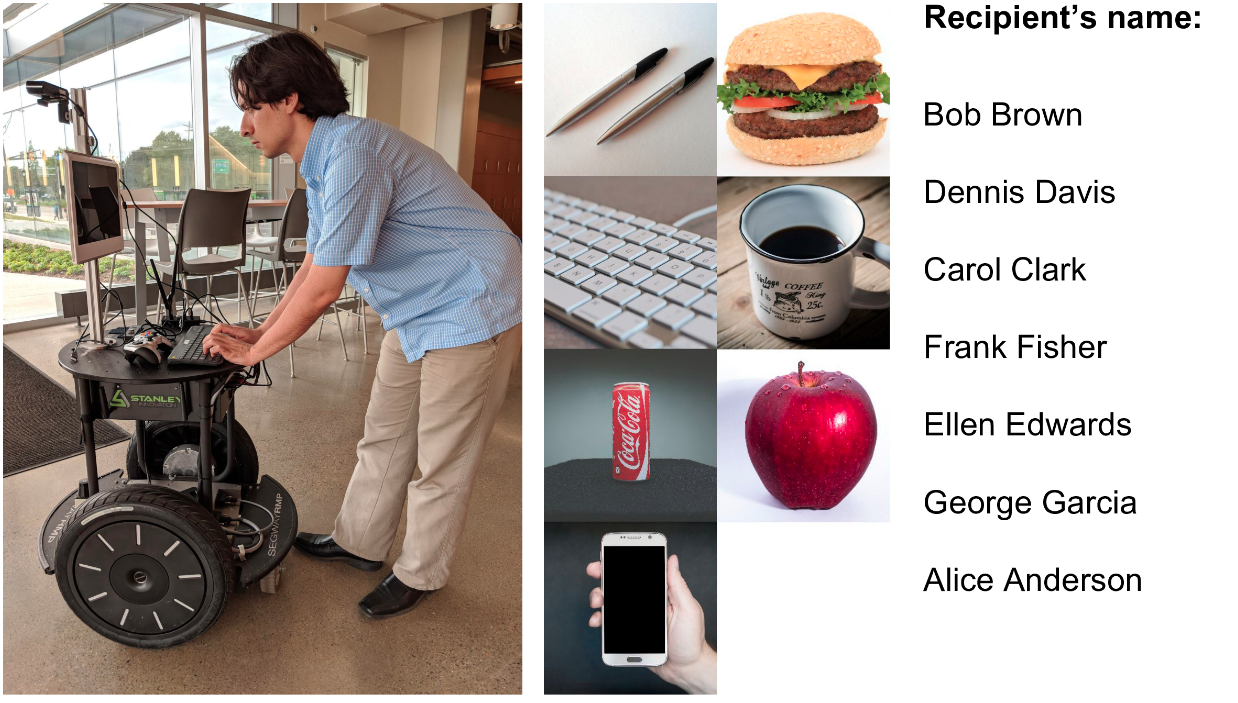}
    \vspace{-.5em}
    \caption{{\it Left}: A user is interacting with our robot. {\it Right}: Items and recipients used for specifying delivery tasks. }
    \label{fig:itemspeople}
  \end{center}
  \vspace{-1em}
\end{figure}

Twelve students of ages 19-30 volunteered to participate in an experiment where they asked the robot to conduct delivery tasks using the items and recipients shown in Fig.~\ref{fig:itemspeople}.
Two items and two recipient in the lists were unknown to the robot, resulting in about $49\%$ (i.e., $1-\frac{5}{7}\times \frac{5}{7}$) of the service requests not requiring knowledge augmentation. 
The participants were not aware of this setting, and  arbitrarily chose any item-recipient pair to form a delivery task. 
Each participant conducted the experiment in two (randomly ordered) trials, where the robot used our dialog agent and a baseline agent with a static KB respectively.  
 
\begin{figure}[t]
  \begin{center}
    \vspace{.5em}
    \hspace*{.30em}
    \includegraphics[width=\columnwidth]{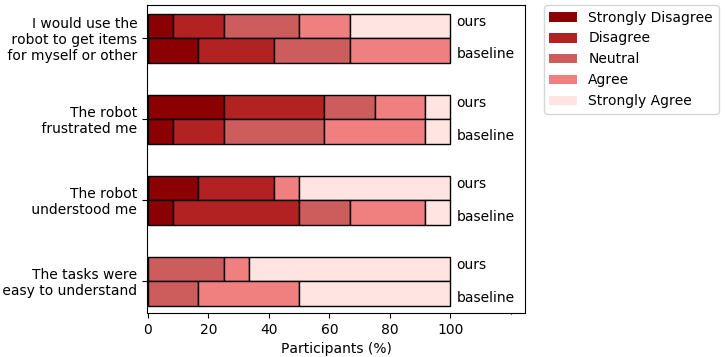}
   \vspace{-1.5em}
    \caption{Results of survey papers from participants, including four statements with Likert-scale responses.  }
    \label{fig:survey}
  \end{center}
  \vspace{-1.5em}
\end{figure}

By the end of each dialog, each participant filled out a survey form that includes prompts: 
Q1, \emph{Task is easy to participants}; Q2, \emph{Robot understood participant}; Q3, \emph{Robot frustrated participant}; Q4, \emph{Participant will use the robot in the future.} 
The response choices range from 0 (Strongly disagree) to 4 (Strongly agree). 
Fig.~\ref{fig:survey} shows results from the survey papers, and Table \ref{tab:1} shows the average scores. 
At the $0.1$ confidence level, our dialog agent performed significantly better in response for Q3 (frustrated) and Q4 (usefulness). 
There is no significance difference observed in responses to the other two questions. 

\paragraph{Mechanical Turk Experiment} 
Experiments have been conducted with 103 human participants via MTurk. 
The setup was same as the robot experiment, except that we used a more challenging baseline. 
The baseline agent augments its KB after a fixed number of $N$ dialog turns ($N=8$ in our case). 
Each worker participated in only one trial (using our agent or the baseline, randomly selected). 
At a confidence level of 0.1, we found our agent to use significantly fewer dialog turns and achieve a significantly higher success rate on average. 
Despite the quantitative improvements, there was no significant difference observed from the scores collected from the survey prompts.\footnote{This could be because the testing domain is static and relatively simple, in the sense of the numbers of items and people. Additionally, MTurk workers are less invested in qualitative feedback than human users in the presence of a real robot.}

\begin{table}[H]\scriptsize
\caption{Results of the human participant experiment.}
\vspace{-1em}
\label{tab:1}
\begin{center}
\begin{tabular}{|l|rrrr|}
 \hline
  & Q1 & Q2 & Q3 & Q4 \\
 \hline
 \hline
 Our dialog agent  & 3.42  & 2.50 & \textbf{1.50}  & \textbf{2.50} \\

 A baseline agent with static KB  & 3.33  & 1.83 & 2.17 &  1.75  \\
 \hline
\end{tabular}
\vspace{-1em}
\end{center}
\end{table}

\paragraph{An Example Trial on a Mobile Robot}

Table \ref{tab:example_dialog} shows the dialog in a human-robot trial.\footnote{The agent was implemented using Robot Operating System (ROS)~\cite{quigley2009ros}.}
The semantic parser failed to correctly parse the request, because it did not know the meaning of \emph{get}. 
As a result, the dialog agent made a random guess that is ``\emph{Alice requesting hamburger}''.
The robot asked again for clarification on what it should deliver.  
The user responded ``\emph{coffee}'', making the robot confused between \emph{coffee} and \emph{hamburger}.  
The robot asked for confirmation about ``\emph{hamburger}'', and got denied by the user.  
Although the user was explicitly guided to specify the recipient by names provided in Figure \ref{fig:itemspeople}, this participant used ``me'', an unknown word to the robot. 
When the robot heard ``\emph{Nate}'' the first time, it could not understand, and mistakenly interpreted it as ``\emph{Ellen}''. 
As a result, the agent confirmed ``\emph{Ellen}'', and got denied. 
After the negative responses, it added (\emph{Nate}) to its KB, and became confident about the request.

\begin{table}[t]\small
\vspace{1em}
\caption{An example dialog from a human participant.}
\vspace{-1em}
\label{tab:example_dialog}
\begin{center}
\begin{tabular}{|L{5.5cm}|L{1.5cm}|}
 \hline
 \textbf{Robot} & \textbf{Human}   \\
 \hline \hline
 How can I help you?  &  Get me coffee  \\
 \hline
  What item should I bring? & Coffee  \\
 \hline
  Do you want me to deliver hamburger? & No  \\
 \hline
 Who should I bring the item to? & Nate  \\
 \hline
Is this delivery for ellen? & No  \\
 \hline
It seems I do not know the person you are talking about.  Please write their name so I can learn it. & Nate  \\
 \hline
Who should I bring the item to? & Nate  \\
 \hline
Is this delivery for Nate?  & Yes \\
 \hline
 What item should I bring? & Coffee \\
 \hline \hline
 
\multicolumn{2}{|l|}{\textbf{Execute: Robot brings coffee for Nate; the dialog is over.}} \tabularnewline
      \hline 
\end{tabular}
\end{center}
\vspace{-1em}
\end{table}

\section{Conclusions \& Future Work}
We introduced a dialog agent that simultaneously supports human intention identification and knowledge augmentation on an as-needed basis. 
Experiments show that our dual-track POMDP controller enables the agent to simultaneously conduct dialog and knowledge management. 
In comparison to a baseline that augments its knowledge base after a fixed number of turns, our dialog agent consistently produces higher overall dialog success. 
Experiments with human participants show that our agent is more successful in augmenting knowledge and estimating human intention, and people are more willing to use our system. In the future, we intend to improve our agent by minimizing the dialog duration (e.g. less double-checking attempts) as well as augmenting its knowledge with more complex structures, e.g., to model subclasses of items. 


\section*{Acknowledgements}
We are grateful to the BWI team at UT Austin for making their software available to the public. Part of this work has taken place in the Autonomous Intelligent Robotics (AIR) group at SUNY Binghamton. AIR research is supported in part by SUNY RF, and Ford.

\addtolength{\textheight}{-12cm}   


{\footnotesize
\bibliographystyle{IEEEtrans}
\bibliography{references}

\begin{thebibliography}{10}
\providecommand{\url}[1]{#1}
\csname url@samestyle\endcsname
\providecommand{\newblock}{\relax}
\providecommand{\bibinfo}[2]{#2}
\providecommand{\BIBentrySTDinterwordspacing}{\spaceskip=0pt\relax}
\providecommand{\BIBentryALTinterwordstretchfactor}{4}
\providecommand{\BIBentryALTinterwordspacing}{\spaceskip=\fontdimen2\font plus
\BIBentryALTinterwordstretchfactor\fontdimen3\font minus
  \fontdimen4\font\relax}
\providecommand{\BIBforeignlanguage}[2]{{%
\expandafter\ifx\csname l@#1\endcsname\relax
\typeout{** WARNING: IEEEtran.bst: No hyphenation pattern has been}%
\typeout{** loaded for the language `#1'. Using the pattern for}%
\typeout{** the default language instead.}%
\else
\language=\csname l@#1\endcsname
\fi
#2}}
\providecommand{\BIBdecl}{\relax}
\BIBdecl

\bibitem{khandelwal2017bwibots}
P.~Khandelwal, S.~Zhang, J.~Sinapov, M.~Leonetti, J.~Thomason, F.~Yang,
  I.~Gori, M.~Svetlik, P.~Khante, V.~Lifschitz \emph{et~al.}, ``{BWIbots}: A
  platform for bridging the gap between {AI} and human--robot interaction
  research,'' \emph{The International Journal of Robotics Research}, 2017.

\bibitem{hawes2017strands}
N.~Hawes, C.~Burbridge, F.~Jovan \emph{et~al.}, ``The strands project:
  Long-term autonomy in everyday environments,'' \emph{IEEE Robotics \&
  Automation Magazine}, vol.~24, no.~3, pp. 146--156, 2017.

\bibitem{chen2017robots}
Y.~Chen, F.~Wu, W.~Shuai, and X.~Chen, ``Robots serve humans in public places
  — kejia robot as a shopping assistant,'' \emph{International Journal of
  Advanced Robotic Systems}, vol.~14, no.~3, 2017.

\bibitem{veloso2018increasingly}
M.~M. Veloso, ``The increasingly fascinating opportunity for human-robot-ai
  interaction: The cobot mobile service robots,'' \emph{ACM Transactions on
  Human-Robot Interaction}, 2018.

\bibitem{thomason2015learning}
J.~Thomason, S.~Zhang, R.~J. Mooney, and P.~Stone, ``Learning to interpret
  natural language commands through human-robot dialog,'' in \emph{Proceedings
  of the 24th International Conference on Artificial Intelligence}, 2015, pp.
  1923--1929.

\bibitem{matuszek2013learning}
C.~Matuszek, E.~Herbst, L.~Zettlemoyer, and D.~Fox, ``Learning to parse natural
  language commands to a robot control system,'' in \emph{Experimental
  Robotics}, 2013, pp. 403--415.

\bibitem{misra2016tell}
D.~K. Misra, J.~Sung, K.~Lee, and A.~Saxena, ``Tell me dave: Context-sensitive
  grounding of natural language to manipulation instructions,'' \emph{The
  International Journal of Robotics Research}, vol.~35, no. 1-3, pp. 281--300,
  2016.

\bibitem{tellex2011understanding}
S.~Tellex, T.~Kollar, S.~Dickerson, M.~R. Walter, A.~G. Banerjee, S.~Teller,
  and N.~Roy, ``Understanding natural language commands for robotic navigation
  and mobile manipulation,'' in \emph{Proceedings of the 25th AAAI Conference},
  2011, pp. 1507--1514.

\bibitem{spranger2015co}
M.~Spranger and L.~Steels, ``Co-acquisition of syntax and semantics: an
  investigation in spatial language,'' in \emph{Proceedings of the 24th
  International Conference on Artificial Intelligence}, 2015.

\bibitem{gong2018temporal}
Z.~Gong and Y.~Zhang, ``Temporal spatial inverse semantics for robots
  communicating with humans,'' in \emph{Proceedings of (ICRA)}, 2018.

\bibitem{singh2002optimizing}
S.~Singh, D.~Litman, M.~Kearns, and M.~Walker, ``Optimizing dialogue management
  with reinforcement learning: Experiments with the njfun system,''
  \emph{JAIR}, vol.~16, pp. 105--133, 2002.

\bibitem{puterman2014markov}
M.~L. Puterman, \emph{Markov decision processes: discrete stochastic dynamic
  programming}.\hskip 1em plus 0.5em minus 0.4em\relax John Wiley \& Sons,
  2014.

\bibitem{sutton1998reinforcement}
R.~S. Sutton and A.~G. Barto, \emph{Reinforcement learning: An
  introduction}.\hskip 1em plus 0.5em minus 0.4em\relax MIT press Cambridge,
  1998.

\bibitem{cuayahuitl2017simpleds}
H.~Cuay{\'a}huitl, ``Simpleds: A simple deep reinforcement learning dialogue
  system,'' in \emph{Dialogues with Social Robots}, 2017.

\bibitem{lu2019goal}
K.~Lu, S.~Zhang, and X.~Chen, ``Goal-oriented dialogue policy learning from
  failures,'' in \emph{AAAI}, 2019.

\bibitem{zhang2015corpp}
S.~Zhang and P.~Stone, ``{CORPP}: commonsense reasoning and probabilistic
  planning, as applied to dialog with a mobile robot,'' in \emph{Proceedings of
  the Twenty-Ninth AAAI Conference on Artificial Intelligence}, 2015, pp.
  1394--1400.

\bibitem{lu2017leveraging}
D.~Lu, S.~Zhang, P.~Stone, and X.~Chen, ``Leveraging commonsense reasoning and
  multimodal perception for robot spoken dialog systems,'' in \emph{IEEE/RSJ
  International Conference on Intelligent Robots and Systems (IROS)}, 2017, pp.
  6582--6588.

\bibitem{padmakumar2017integrated}
A.~Padmakumar, J.~Thomason, and R.~J. Mooney, ``Integrated learning of dialog
  strategies and semantic parsing,'' in \emph{Proceedings of the 15th
  Conference of the European Chapter of the ACL}, 2017, pp. 547--557.

\bibitem{mericcli2014interactive}
c.~Meri{\c{c}}li, S.~D. Klee, J.~Paparian, and M.~Veloso, ``An interactive
  approach for situated task specification through verbal instructions,'' in
  \emph{Proceedings of the 2014 international AAMAS conference}, 2014.

\bibitem{perera2015learning}
V.~Perera, R.~Soetens, T.~Kollar, M.~Samadi, Y.~Sun, D.~Nardi, R.~van~de
  Molengraft, and M.~Veloso, ``Learning task knowledge from dialog and web
  access,'' \emph{Robotics}, vol.~4, no.~2, pp. 223--252, 2015.

\bibitem{azaria2016instructable}
A.~Azaria, J.~Krishnamurthy, and T.~M. Mitchell, ``Instructable intelligent
  personal agent.'' in \emph{AAAI}, 2016, pp. 2681--2689.

\bibitem{tucker2017learning}
M.~Tucker, D.~Aksaray, R.~Paul, G.~J. Stein, and N.~Roy, ``Learning unknown
  groundings for natural language interaction with mobile robots,'' in
  \emph{ISRR}, 2017.

\bibitem{she2017interactive}
L.~She and J.~Chai, ``Interactive learning of grounded verb semantics towards
  human-robot communication,'' in \emph{Proceedings of ACL}, 2017, pp.
  1634--1644.

\bibitem{chai2018language}
J.~Y. Chai, Q.~Gao, L.~She, S.~Yang, S.~Saba-Sadiya, and G.~Xu, ``Language to
  action: Towards interactive task learning with physical agents.'' in
  \emph{IJCAI}, 2018, pp. 2--9.

\bibitem{jiang2019open}
Y.~Jiang, N.~Walker, J.~Hart, and P.~Stone, ``Open-world reasoning for service
  robots,'' in \emph{Proceedings of the 29th International Conference on
  Automated Planning and Scheduling (ICAPS)}, 2019.

\bibitem{kaelbling1998planning}
L.~P. Kaelbling, M.~L. Littman, and A.~R. Cassandra, ``Planning and acting in
  partially observable stochastic domains,'' \emph{Artificial intelligence},
  vol. 101, no. 1-2, pp. 99--134, 1998.

\bibitem{young2013pomdp}
S.~Young, M.~Ga{\v{s}}i{\'c}, B.~Thomson, and J.~D. Williams, ``{POMDP}-based
  statistical spoken dialog systems: A review,'' \emph{Proceedings of the
  IEEE}, vol. 101, no.~5, pp. 1160--1179, 2013.

\bibitem{artzi2013cornell}
Y.~Artzi, ``Cornell {SPF}: Cornell semantic parsing framework,'' \emph{arXiv
  preprint arXiv:1311.3011}, 2013.

\bibitem{gelfond2014knowledge}
M.~Gelfond and Y.~Kahl, \emph{Knowledge representation, reasoning, and the
  design of intelligent agents: The answer-set programming approach}.\hskip 1em
  plus 0.5em minus 0.4em\relax Cambridge University Press, 2014.

\bibitem{kurniawati2008sarsop}
H.~Kurniawati, D.~Hsu, and W.~S. Lee, ``Sarsop: Efficient point-based pomdp
  planning by approximating optimally reachable belief spaces.'' in \emph{RSS},
  vol. 2008.\hskip 1em plus 0.5em minus 0.4em\relax Zurich, Switzerland., 2008.

\bibitem{quigley2009ros}
M.~Quigley, K.~Conley, B.~Gerkey, J.~Faust, T.~Foote, J.~Leibs, R.~Wheeler, and
  A.~Y. Ng, ``Ros: an open-source robot operating system,'' in \emph{ICRA
  workshop on open source software}, 2009.

\end{thebibliography}
}

\end{document}